\documentclass[conference]{IEEEtran}
\IEEEoverridecommandlockouts

\usepackage{cite}
\usepackage{amsmath,amssymb,amsfonts}
\usepackage{graphicx}
\usepackage{textcomp}
\usepackage{xcolor}
\usepackage{booktabs}
\usepackage{multirow}
\usepackage{colortbl}

\begin{document}

\title{VISTA: Video-Injected Stylized Text-to-Animation}

\author{
\IEEEauthorblockN{Monseej Purkayastha\textsuperscript{1},
Anindita Ghosh\textsuperscript{1,2},
Philipp Slusallek\textsuperscript{1}}\\[4pt]
\IEEEauthorblockA{\textsuperscript{1}Saarland Informatics Campus \& German Research Centre for Artificial Intelligence
(DFKI),
Saarbr\"{u}cken, Germany}\\
\IEEEauthorblockA{\textsuperscript{2}Max Planck Institute for Informatics (MPII), Saarbr\"{u}cken, Germany}\\[2pt]
\IEEEauthorblockA{\{monseej.purkayastha, anindita.ghosh, philipp.slusallek\}@dfki.de}
}

\maketitle

\begin{abstract}
We present VISTA, a two-stage framework for generating stylized 3D human motion by fusing structural content from text prompts with expressive style from reference videos, without requiring jointly paired (text, video, stylized motion) triplets. A Dual-channel Autoencoder first maps motion sequences and video clips into a shared latent manifold. A masked autoregressive diffusion backbone then operates within this manifold, injecting video-derived style through a dedicated late-fusion Dual-AdaLN pathway while preserving text-conditioned content structure. A cross-batch unpaired training protocol with latent cycle consistency enables joint learning across separate semantically rich and stylistically diverse datasets. As a proof-of-concept for controllable animation synthesis, we validate VISTA on rendered motion-capture references: it achieves the highest style recognition accuracy among video-conditioned methods while preserving competitive content alignment, and its decomposed 3-way classifier-free guidance provides independent, user-controllable calibration of the content--style balance at inference time.
\end{abstract}


\section{Introduction}
Generating 3D human motion from natural language has advanced rapidly through diffusion-based approaches~\cite{mdm, mld, mardm}, yet extending these models to synthesize stylized motion from multimodal inputs, for instance, combining text prompts with video style references introduces two key challenges. First, standard encoders like CLIP~\cite{clip} process video frames independently, discarding the causal inter-frame structure that distinguishes one motion style from another. Second, no dataset provides paired (text, video, stylized motion) triplets: content datasets like HumanML3D~\cite{humanml3d} offer text--motion pairs, while style datasets like 100STYLES~\cite{100styles} provide labelled stylized motion without video.

Prior methods partially address these gaps. SMooDi~\cite{smoodi} uses a ControlNet-style~\cite{controlnet} adaptor with decomposed CFG, LoRA-MDM~\cite{loramdm} applies low-rank adaptation of attention weights, and StyleMotif~\cite{stylemotif} introduces style-content cross-normalization. However, these methods either operate on motion references rather than RGB video, or struggle to preserve style during dynamic action phases due to modality collision~\cite{smoodi, stylemotif}.

We propose VISTA (Video-Injected Stylized Text-to-Animation), which targets both challenges through: (1)~a shared latent manifold bridging video and motion domains via a Dual-channel Autoencoder, (2)~a late-fusion diffusion backbone with dedicated style blocks and Dual-AdaLN modulation maintaining strict content--style separation, and (3)~a cross-batch unpaired training protocol with latent cycle consistency.

\section{Preliminaries}
\noindent 
\textbf{Motion Representation.}
Following Meng et al.~\cite{mardm}, we represent motion as a 67-dimensional per-frame feature vector (root trajectory + 21 root-relative joint positions), a compact subset of the 263-dimensional RIC representation~\cite{humanml3d}.

\noindent
\textbf{Video Representation.}
We require paired motion-video annotations to train the video encoder. For this, we render the corresponding motion sequences from the 100STYLES dataset~\cite{100styles} as an SMPL~\cite{smpl} mesh (VE-100STYLES). The renders are taken from the front and the left view of the character. We process these videos through a frozen ViViT-B/16$\times$2~\cite{vivit} backbone, which extracts spatio-temporal tubelets producing feature tensors.

\noindent \textbf{Text Representation.} 
We encode text prompts via a frozen CLIP~\cite{clip} text encoder.

\section{Method}

\begin{figure}[t]
    \centering
    \includegraphics[width=1\linewidth]{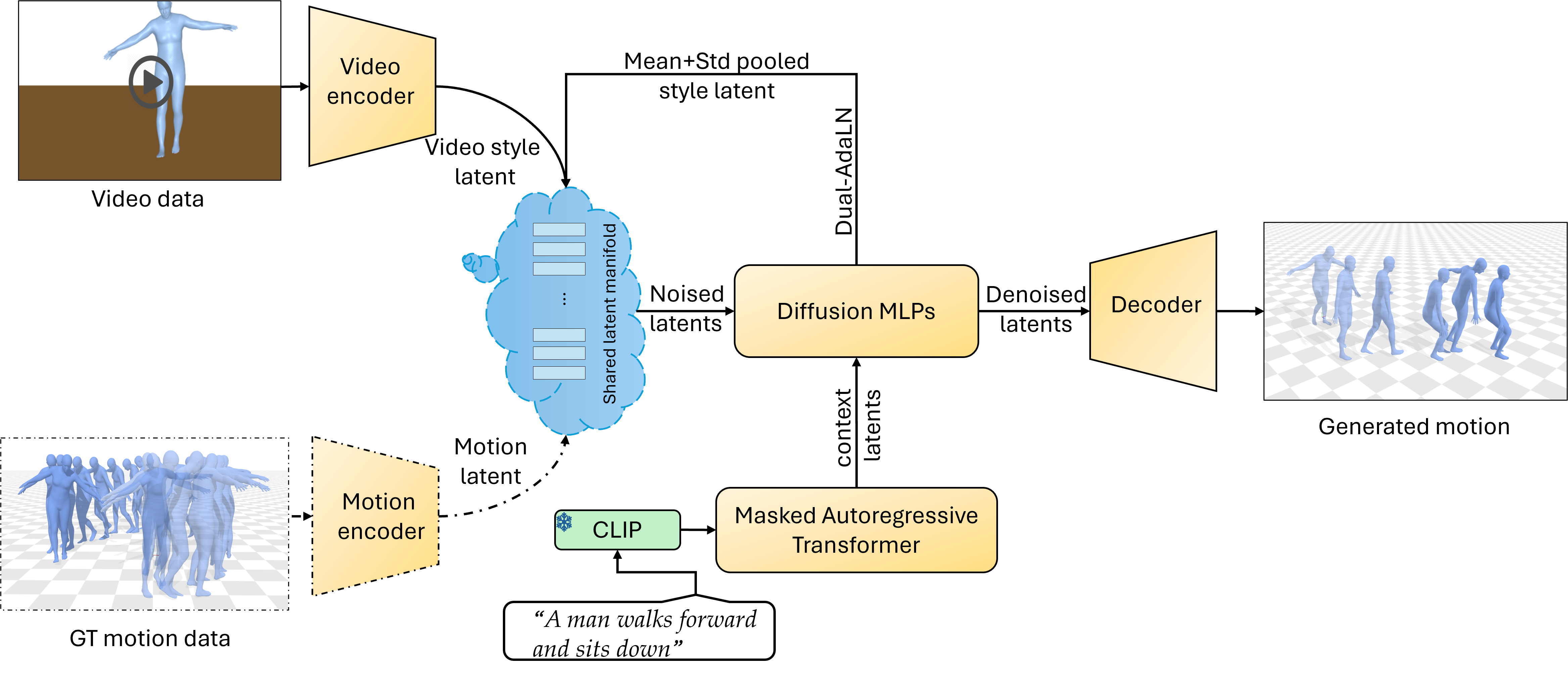}
    \caption{Overview of the VISTA framework. Video and motion encoders map into a shared latent manifold; text conditions the MARTransformer via CLIP; style is injected into the Diffusion MLPs via Dual-AdaLN. Dotted components are used during training only.}
    \label{fig:overview}
\end{figure}

\textbf{Stage~1: Dual-Channel Autoencoder.}
The DualAE compresses motion sequences and rendered video clips into a shared latent manifold. A motion encoder and a video encoder, built atop the frozen ViViT backbone produce aligned latent codes. A shared decoder reconstructs motion from either latent, trained with reconstruction, embedding alignment (with stop-gradient on the motion-side), style classification, adversarial, and cycle-consistency losses in a phased curriculum.

\textbf{Temporal Style Extraction.}
We apply concatenated mean-and-standard-deviation pooling over the video latent sequence, preserving the causal temporal variance that mean pooling collapses. The resulting vector is projected via a two-layer MLP into a 512-dimensional style token. Crucially, by discarding frame-ordered structure, this temporal pooling encourages the video branch toward stylistic rather than action-level modulation, reducing the risk that frame-level action content from the reference video leaks into the diffusion backbone.

\textbf{Stage~2: Late-Fusion Diffusion.}
Building on a masked autoregressive diffusion backbone (MARDM~\cite{mardm}), VISTA routes text and style through architecturally separated pathways. A text-conditioned MARTransformer (bidirectional self-attention blocks) produces position-specific context vectors for masked latent positions. A Diffusion MLP (DiffMLP) stack of content blocks followed by dedicated style blocks denoises these positions using Dual-AdaLN: each block regresses shift, scale, and gate parameters from both context and style vectors, fused via additive (shift, gate) and bounded multiplicative (scale) combination with a block-wise linear weight schedule that progressively amplifies style influence through network depth. We zero-initialize style projections for identity-at-init stability.

\textbf{Cross-Batch Unpaired Training.}
Three concurrent passes per iteration bridge the domain gap. Pass~1 trains on HumanML3D text--motion pairs (content preservation). Pass~2 trains on VE-100STYLES with both text and video conditioning. Pass~3 (activated with 50\% probability) performs an unpaired cycle: a content latent is styled with an unpaired video reference, then the styled output is detached and passed back through the model with null style to recover the original content, supervised by masked-position cycle consistency. Independent CFG dropout (5\% both, 10\% text-only, 10\% style-only) enables the decomposed guidance formulations detailed in the next section.

\textbf{Inference.}
At inference, VISTA iteratively unmasks a fully masked latent sequence via a cosine schedule. The independent condition dropout during training enables two classifier-free guidance (CFG) modes: (1)~2-way CFG runs a pass conditioned on both text and style and an unconditional pass, blended with a single scale $s$. (2)~3-way CFG decomposes guidance into three passes: text-only, style-only, and unconditional, with independent scales $s_{style}, s_{text}$ respectively for each, giving direct user control over the content--style balance without retraining.

\section{Experiments}

We evaluate on 5 styles from VE-100STYLES (Aeroplane, ArmsFolded, Chicken, Robot, Superman) across three settings: base text-to-motion generation on HumanML3D, in-domain styled generation from VE-100STYLES, and zero-shot cross-modal transfer between the two datasets. We compare against SMooDi~\cite{smoodi} and LoRA-MDM~\cite{loramdm}, evaluating all methods using our shared conditioning pipeline on identical splits. Our evaluation uses rendered videos from motion-capture data under controlled conditions (fixed cameras, clean backgrounds, neutral body shape), and we position VISTA as a preliminary study in such controlled settings.

\begin{table}[t]
\centering
\caption{Styled generation and zero-shot cross-modal transfer results. SRA$_1$: Top-1 style recognition accuracy; R-Prec$_3$: Top-3 R-Precision. VISTA-2way uses joint guidance ($s{=}4.5$); VISTA-3way decomposes into $s_{text}{=}4.5$, $s_{style}{=}2.0$.}
\label{tab:results}
\vspace{2pt}
\resizebox{\columnwidth}{!}{%
\begin{tabular}{l cc cc c}
\toprule
 & \multicolumn{2}{c}{\textbf{Styled}} & \multicolumn{2}{c}{\textbf{Transfer}} & \textbf{Base} \\
\cmidrule(lr){2-3} \cmidrule(lr){4-5} \cmidrule(lr){6-6}
\textbf{Method} & SRA$_1$ $\uparrow$ & FID $\downarrow$ & SRA$_1$ $\uparrow$ & R-P$_3$ $\uparrow$ & FID $\downarrow$ \\
\midrule
SMooDi~\cite{smoodi} & 68.1 & 9.74 & 48.0 & \textbf{75.0} & \textbf{0.80} \\
LoRA-MDM~\cite{loramdm} & 55.3 & 12.67 & 14.0 & 60.4 & 4.42 \\
VISTA-2way & \textbf{77.8} & 4.74 & \textbf{70.0} & 60.4 & 2.49 \\
VISTA-3way & 59.7 & \textbf{4.27} & 50.0 & 68.8 & 2.49 \\
\bottomrule
\end{tabular}%
}
\end{table}

We evaluate using standard metrics for stylized motion synthesis: Fr\'{e}chet Inception Distance (FID)~\cite{fid} for distributional quality, Style Recognition Accuracy (SRA)~\cite{smoodi} for stylistic fidelity, and R-Precision~\cite{humanml3d} for text--motion semantic alignment. To avoid train--evaluation coupling, SRA is computed with an independent classifier trained only on ground-truth motion, architecturally and parametrically distinct from the network used in the training style loss.

Table~\ref{tab:results} summarizes the key results. In styled generation, VISTA-2way achieves the highest style recognition accuracy and substantially lower FID. In zero-shot transfer, where generated motions have no ground-truth pairing, VISTA-2way achieves a higher SRA$_1$ compared to SMooDi. SMooDi achieves higher content alignment (R-Prec$_3$), consistent with its conservative residual design. This content--style tradeoff is inherent to the evaluation: R-Precision is computed against neutral (unstyled) text--motion pairs, so higher R-Precision in the styled setting indicates greater style suppression rather than better semantic understanding, as confirmed by the ground-truth R-Prec$_3$ of only 53.1\%. VISTA-3way occupies the middle ground, confirming that the decomposed guidance provides user-controllable calibration of the content--style tradeoff without retraining. We report VISTA-3way at $s_{style}{=}2.0$ rather than a higher scale because this operating point minimizes styled FID while preserving output diversity near the ground-truth level (6.85 vs.\ GT 7.06); users seeking stronger style can increase $s_{style}$ at the cost of diversity, as quantified in Table~\ref{tab:cfg_sweep}.

We also evaluate VISTA on a style-content conflict, pairing ``A person claps with both hands while walking forward" with Aeroplane style. The output reveals direct competition between the two signals: content dominates in most frames, with the subject walking forward and performing clapping motions; only in the final few frames does the style come out. Additionally, we evaluate the model on a single YouTube video showing Aeroplane-style movement with the prompt "A person walks forward and sits down." The output adheres to neither signal, producing a circular running motion in Superman style. This failure likely stems from the ViViT encoder being unable to extract a meaningful style representation from unconstrained video, and the resulting noisy style vector interfering with the content pathway during Dual-AdaLN injection.

\subsection{Baseline Adaptation to Video Conditioning}
Neither SMooDi nor LoRA-MDM was originally designed for video-conditioned style transfer. SMooDi conditions on a reference motion sequence, while LoRA-MDM encodes style through per-style text tokens and per-style LoRA weights. To enable comparison, both baselines are adapted to consume video-derived features with minimal architectural changes. All three models receive features from the same frozen ViViT-B/16$\times$2 backbone, pre-computed offline.

\textbf{SMooDi.} We add a single learned projection that maps post-VideoEncoder latents from VISTA's trained DualAE into the style encoder's input dimension. The style encoder's motion-specific input projection is bypassed, and the style encoder transformer is fully retrained. The ControlNet injection mechanism and frozen MLD backbone remain unchanged. 

\textbf{LoRA-MDM.} We add a set of projection layers that map the raw ViViT embeddings into the transformer's latent space as prefix tokens. The frozen MDM backbone and rank-5 LoRA adapters are retained. A three-phase training curriculum (text-only $\rightarrow$ mixed $\rightarrow$ video-only) prevents the model from ignoring video signals early in training.

\subsection{Ablation Studies}
\textbf{Guidance Scale Sensitivity.}
We sweep $s_{style}$ from 1.0 to 5.0 with $s_{text} = 4.5$ fixed (Table~\ref{tab:cfg_sweep}). In styled generation, top-1 SRA rises monotonically from 38.9\% to 75.0\%, while R-Precision degrades from 93.8\% to 70.3\%, confirming the content--style tradeoff. FID reaches its optimum at $s_{style} = 2.0$ before rising. In transfer, SRA$_1$ peaks at $s_{style} \in [4.0, 4.5]$ (70.0\%) and declines at higher values, indicating that excessive style guidance degrades transfer quality.

\begin{table}[t]
\centering
\caption{Guidance scale sensitivity: effect of $s_{style}$ ($s_{text}{=}4.5$ fixed) on styled generation and zero-shot transfer metrics.}
\label{tab:cfg_sweep}
\vspace{2pt}
\resizebox{\columnwidth}{!}{%
\begin{tabular}{l ccc ccc}
\toprule
 & \multicolumn{3}{c}{\textbf{Styled}} & \multicolumn{3}{c}{\textbf{Transfer}} \\
\cmidrule(lr){2-4} \cmidrule(lr){5-7}
$s_{style}$ & SRA$_1$  $\uparrow$& R-P$_3$  $\uparrow$ & FID $\downarrow$ & SRA$_1$  $\uparrow$ & R-P$_3$  $\uparrow$ & Skt. $\downarrow$\\
\midrule
1.0 & 38.9 & 93.8 & 6.53 & 30.0 & 87.5 & 0.11 \\
2.0 & 59.7 & 85.9 & \textbf{4.27} & 50.0 & 68.8 & 0.14 \\
3.0 & 66.7 & 82.8 & 5.21 & 62.0 & 64.6 & 0.16 \\
4.0 & 70.8 & 76.6 & 5.87 & \textbf{70.0} & 60.4 & 0.17 \\
4.5 & 73.6 & 73.4 & 5.85 & \textbf{70.0} & 50.0 & 0.18 \\
5.0 & \textbf{75.0} & 70.3 & 5.80 & 68.0 & 54.2 & 0.19 \\
\bottomrule
\end{tabular}%
}
\end{table}

\section{Conclusion and Future Work}

VISTA demonstrates that video-conditioned stylized motion generation is viable without jointly paired multimodal ground truth, achieving the highest style recognition among video-conditioned methods through late-fusion routing and cross-batch cycle consistency, with decomposed 3-way CFG enabling precise content--style calibration at inference time.

Three directions follow: First, replacing the discrete classification objective in Stage~1 with a contrastive loss would encourage a continuous style manifold, enabling interpolation, composition, and blending of styles beyond the training set. Second, while the current evaluation focuses on rendered motion-capture video in a controlled setting, extending the video encoder to handle unconstrained in-the-wild video with occlusions, dynamic cameras, and visual clutter is the primary next step toward practical applicability; preliminary experiments with real-world video references are under way. Third, replacing the static global style vector with token-level cross-attention over the full video latent sequence would enable per-frame style modulation, directly addressing the conditioning pathway asymmetry that causes style fading during dynamic action phases.

\section{Acknowledgment}
This work was done in the context of the IntelSaar-Animations, Future of Graphics and Media: Avatar Latency Compensation project and has been funded by the German State of Saarland and Intel Corporation (GRA 5032 - 05039) and the German Federal Ministry for Economic Affairs and Energy (BMWE) in the TwinMap project (13IK028J).


\bibliographystyle{IEEEtran}

\end{document}